\documentclass[a4paper, 10pt, conference]{ieeeconf}
\IEEEoverridecommandlockouts

\usepackage[
    left=54pt,
    right=37pt,
    top=104pt,
    bottom=54pt
]{geometry}

\usepackage{bm}
\usepackage{url}
\usepackage{cite}
\usepackage{tikz}
\usepackage{xfrac}
\usepackage{xargs}
\usepackage{times}
\usepackage{array}
\usepackage{siunitx}
\usepackage{balance}
\usepackage{relsize}
\usepackage{amssymb}
\usepackage{amsmath}
\usepackage{amsfonts}
\usepackage{mathrsfs}
\usepackage{scalerel}
\usepackage{stfloats}
\usepackage{flushend}
\usepackage{graphics}
\usepackage{graphicx}
\usepackage{hyperref}
\usepackage{multicol}
\usepackage{verbatim}
\usepackage{multirow}
\usepackage{booktabs}
\usepackage{pgfplots}
\usepackage{stfloats}
\usepackage{textcomp}
\usepackage{algorithm}
\usepackage{xinttools}
\usepackage{glossaries}
\usepackage{algorithmic}
\usepackage[nolist]{acronym}
\usepackage[dvipsnames]{xcolor}

\usepackage[caption=false,font=normalsize,labelfont=sf,textfont=sf]{subfig}

\usetikzlibrary{arrows.meta}
\usepgfplotslibrary{statistics}
\usepgfplotslibrary{groupplots}
\usepgfplotslibrary{fillbetween}

\pgfplotsset{compat=1.16}
\graphicspath{{figures/}}
\DeclareGraphicsExtensions{.pdf,.png,.jpg,.jpeg}

\definecolor{wong_gray}{HTML}{888888}%
\definecolor{wong_black}{HTML}{333333}%
\definecolor{wong_gold}{HTML}{E69F00}%
\definecolor{wong_cyan}{HTML}{56B4E9}%
\definecolor{wong_green}{HTML}{009E73}%
\definecolor{wong_yellow}{HTML}{F0E442}%
\definecolor{wong_blue}{HTML}{0072B2}%
\definecolor{wong_red}{HTML}{D55E00}%
\definecolor{wong_pink}{HTML}{CC79A7}%
\definecolor{wong_magenta}{HTML}{CA1963}%

\definecolor{skin}{HTML}{FF9999}%
\definecolor{antiskin}{HTML}{005555}%
\definecolor{shoe}{HTML}{553311}%
\definecolor{tiago}{HTML}{AAAAAA}%
\definecolor{model_cyan}{HTML}{87D2FF}%
\definecolor{model_purple}{HTML}{BA55D3}%
\definecolor{model_yellow}{HTML}{FFAA00}%
\definecolor{kinect}{HTML}{FF0000}%
\definecolor{lidar}{HTML}{11CC22}%
\definecolor{optitrack}{HTML}{020EBB}%

\newacro{fov}[FOV]{Field of View}
\newacro{fcn}[FCN]{Fully Convolutional Network}
\newacro{hri}[HRI]{Human-Robot Interaction}
\newacro{fcn}[FCN]{Fully Convolutional Network}
\newacro{nms}[NMS]{Non-maxima Suppression}

\title{\LARGE \bf
Sixth-Sense: Self-Supervised Learning\\of Spatial Awareness of Humans from a Planar Lidar
}

\author{Simone Arreghini, Nicholas Carlotti, Mirko Nava, Antonio Paolillo, and Alessandro Giusti%
\thanks{All authors are with the Dalle Molle Institute for Artificial Intelligence (IDSIA), USI-SUPSI, Lugano, Switzerland. Corresponding author: Simone Arreghini, \texttt{simone.arreghini@idsia.ch}}%
\thanks{This work was supported by the European Union through the project SERMAS, by the Swiss State Secretariat for Education, Research and Innovation (SERI) under contract number 22.00247, and by the Swiss National Science Foundation, grant number 213074.}%
}%

\begin{document}

\maketitle

\begin{tikzpicture}[remember picture,overlay]
\node[
    anchor=south west,
    inner sep=0pt,
    outer sep=0pt
] at ([xshift=54pt,yshift=6mm]current page.south west) {%
    \fbox{%
        \parbox{\dimexpr\textwidth-2\fboxsep-2\fboxrule\relax}{%
            \footnotesize
            \copyright~2026 IEEE. Personal use of this material is permitted.
            Permission from IEEE must be obtained for all other uses, in any
            current or future media, including reprinting/republishing this
            material for advertising or promotional purposes, creating new
            collective works, for resale or redistribution to servers or lists,
            or reuse of any copyrighted component of this work in other works.
        }%
    }%
};
\end{tikzpicture}

\begin{abstract}
Reliable localization of people is fundamental for service and social robots that must operate in close interaction with humans. State-of-the-art human detectors often rely on RGB-D cameras or costly 3D LiDARs. However, most commercial robots are equipped with cameras with a narrow field of view, leaving them unaware of users approaching from other directions, or inexpensive 1D LiDARs whose readings are hard to interpret. To address these limitations, we propose a self-supervised approach to detect humans and estimate their 2D pose from 1D LiDAR data, using detections from an RGB-D camera as supervision. Trained on 70 minutes of autonomously collected data, our model detects humans omnidirectionally in unseen environments with 71\% precision, 80\% recall, and mean absolute errors of 13~cm in distance and 44$\bm{^\circ}$ in orientation, measured against ground truth data. Beyond raw detection accuracy, this capability is relevant for robots operating in shared public spaces, where omnidirectional awareness of nearby people is crucial for safe navigation, appropriate approach behavior, and timely human-robot interaction initiation using low-cost, privacy-preserving sensing. Deployment in two additional public environments further suggests that the approach can serve as a practical wide-FOV awareness layer for socially aware service robotics.
\end{abstract}


\section{Introduction}\label{sec:intro}
Social and service robots operating in human-populated environments~\cite{Benitti:cae:2012, Gonzalez:as:2021, Choi:jhmm:2020} must be capable to perceive humans, predict their behavior~\cite{Zaraki:icrm:2014} and interaction intention~\cite{Abbate:ras:2024,Arreghini:icra:2024}.
Humans' detection and localization are crucial for \acp{hri} where humans and robots are close, to ensure safe cooperation, improve navigation in crowded spaces, and provide insightful interaction-relevant spatial cues. 
In service and assistance scenarios, knowing where people are around the robot and whether they are oriented toward it is not only useful for perception, but also for shaping the robot social behavior, such as deciding when to yield, when to turn and attend, and when an approach is most likely to be socially appropriate rather than intrusive.
A typical sensing setup for service robots includes a combination of a wide \ac{fov} laser sensor, often seeing $360^{\circ}$ around the robot, and a narrow \ac{fov} RGB-D camera. 
Although 3D LiDAR sensors offer rich environmental information~\cite{martin2021jrdb}, their use in service robots is limited to only few high-end platforms.
Instead, the majority uses simpler 1D planar LiDAR sensors, typically located at wheel height, balancing production costs and richness of information.
Examples span from large research robots~\cite{pages2016tiago} to those commonly present in many households such as robot vacuum cleaners, or lawnmowers~\cite{mahdi2022survey}.
In this context, a $360^{\circ}$ perception system could enable robust human-aware sensing with cost-effective hardware, supporting in-the-wild service applications~\cite{Arreghini:iros:2024} by extending capabilities to mobile robots.

Deep learning models capable of reliably detecting and localizing humans from RGB-D data have been extensively investigated in literature~\cite{paul2013human,tolgyessy2021skeleton}.
However, doing the same accurately with 1D laser sensors remains a challenge due to the sparse nature and complex interpretation of the sensor's readings.
Notably, the typical environment for a service robot may feature structures that result in readings mimicking the human profile, such as the legs of tables, bars in railings, or pets.
Differentiating humans from the rest requires recognizing subtle geometric and dynamic patterns.
Nonetheless, awareness of the presence and direction of nearby humans significantly improves the robot's behavior in social contexts~\cite{dhiman2015creating}, even when these perceptions are uncertain and inaccurate.
Indeed, inspired by animal perception where peripheral vision and hearing direct visual focus towards areas of interest~\cite{vernon1933peripheral}, a robot could use uncertain detections from 1D LiDAR data to trigger further sensing by the more reliable RGB-D camera.
\begin{figure}[t]%
    \centering%
    \frame{\includegraphics[width=1.0\linewidth]{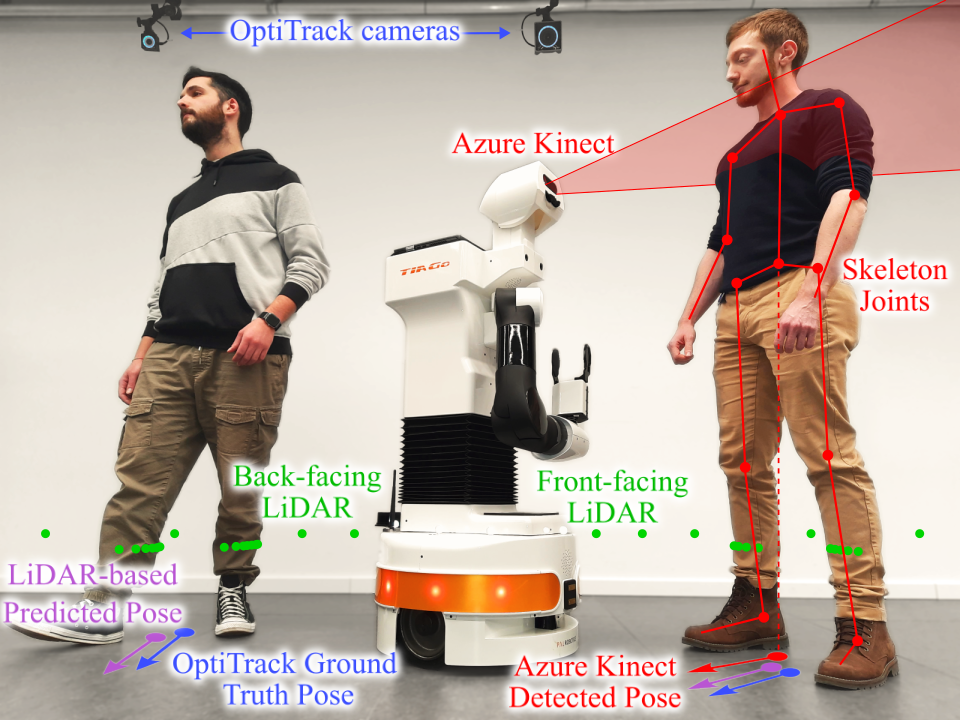}}%
    \caption{Our approach uses a human detector based on the narrow FOV {\color{kinect}Azure Kinect} as a source of labels to train a {\color{model_purple!75!black}1D \ac{fcn}} that, given planar {\color{lidar!75!black}LiDAR} scans, predicts the presence and relative 2D pose of humans around the robot. The training approach relies only on hardware onboard the robot and can autonomously collect data in any environment. 
    When available, a {\color{optitrack}Motion Capture} system collects ground truth used for evaluation purposes.}%
    \label{fig:task_and_sensor_setup}%
    \vspace{-5mm}
\end{figure}
Unlike previous approaches that rely on 3D LiDAR~\cite{yan2020online, hayton2020cnn}, we handle a less informative sensing modality, leading to inherently lower detection accuracy but with a potential widespread adoption by many robotic platforms. 

Our approach relies on a deep learning model that can be trained, or fine-tuned, directly by the robot during its deployment using self-supervision: 
an off-the-shelf detector receiving data from the onboard RGB-D camera is used to provide detections of humans considered as training labels, see Fig.~\ref{fig:task_and_sensor_setup}.
This is an instance of the class of approaches using one sensor to supervise the training of a model which interprets data from a different sensor~\cite{nava2019learning, nava2021uncertainty}; the same paradigm has been applied to skeleton joint pose estimation from 3D LiDAR, using an image-based human detector providing the 2D skeleton joints pose as supervision~\cite{cong2023weakly}.

Our model is fed with a moving time window of 1D LiDAR scans and predicts the presence, 2D position and relative bearing of humans around the robot.  
We employ a loss function to minimize the distance between the model's predicted detections from 1D LiDAR and those coming from the detector, enforced only where the two \acp{fov} overlap.
In this context, the self-supervised learning paradigm enables the model to automatically adapt to its deployment environment and sensor characteristics, removes the need for large pre-collected datasets, and increases robustness to cluttered scenes that may produce human-like false readings.
Further, we adopt a 1D \ac{fcn}~\cite{long2015fully} and leverage its translation-invariance to extend the detection ability to the wider \ac{fov} of the LiDAR, even in directions never covered by the camera.

We describe our \textbf{main contribution} in Sec.~\ref{sec:method}: a practical methodology and open-source implementation for training and running a human detector and pose estimator from 1D LiDAR data using camera detections as self-supervision.
As a further contribution, we publish the datasets collected using this setup, and the pre-trained models resulting from our approach. 
More generally, we frame our work not only as a LiDAR perception method, but also as an enabling component for human-aware robot behavior on platforms that cannot rely on omnidirectional RGB-D or 3D LiDAR sensing.
The remainder of the paper is organized as follows: Sec.~\ref{sec:setup} describes the experimental setup; Sec.~\ref{sec:results} reports the quantitative evaluation against ground-truth data, including validation in previously unseen public environments; Sec.~\ref{sec:conclusions} discusses final remarks and future research directions.
\section{Methodology}\label{sec:method}
We train a 1D \ac{fcn} model on the task of estimating 2D poses of humans around the robot, giving as input a time window of readings from a planar LiDAR sensor located at the center of the robot base with a uniform angular resolution across the entire \ac{fov}.
\begin{figure}[b]
\vspace{-5mm}
\centering%
\frame{\includegraphics[trim={0cm 0.75cm 0cm 0cm}, clip=true,width=\columnwidth]{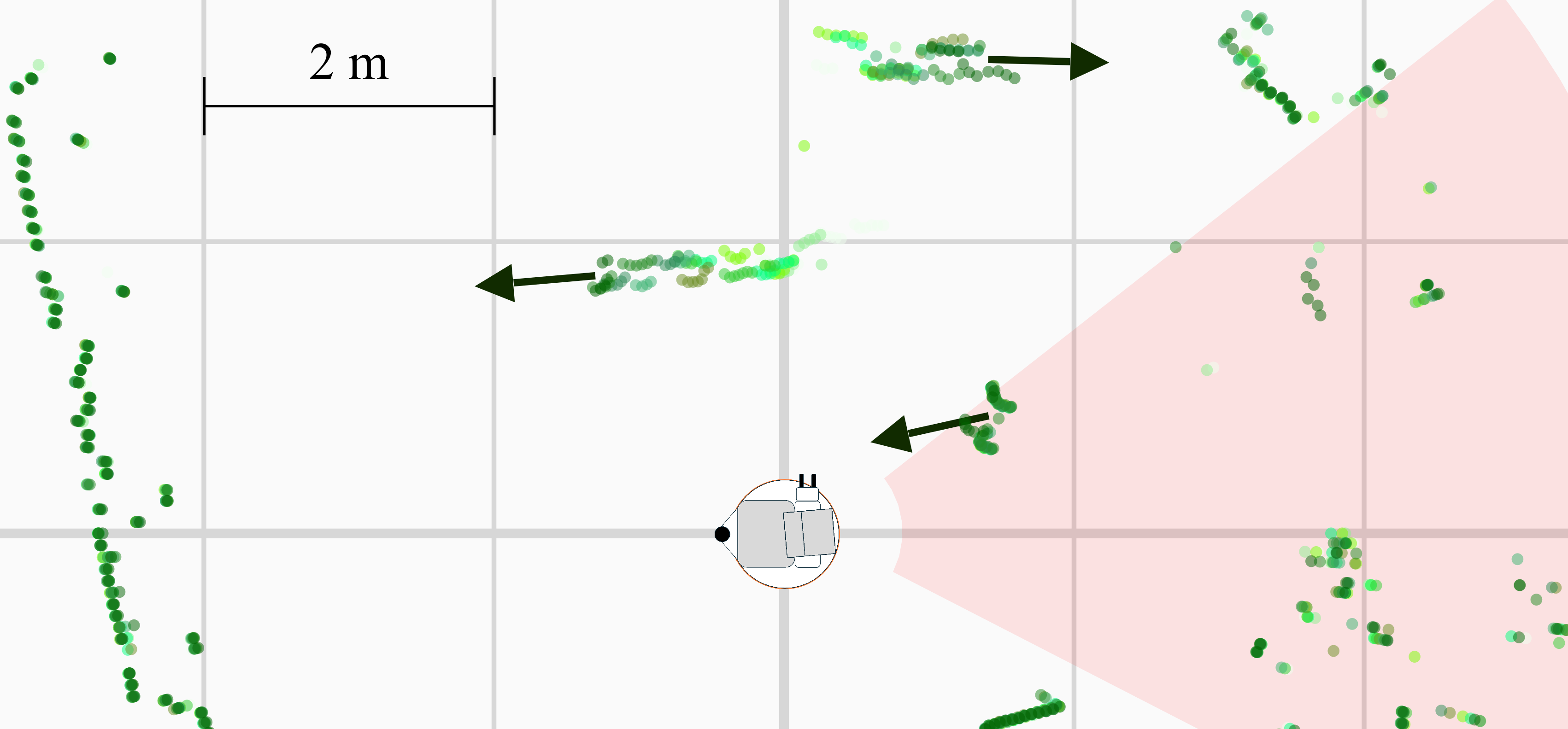}}%
\caption{Walking people and static structures as perceived by the LiDAR: lighter shades of green indicate older scans in the temporal window, whereas black arrows indicate the people's instantaneous orientation.}%
\label{fig:history}%
\end{figure} 
In practice, we use a tensor of $360$ elements representing rays equally spaced around the robot with a resulting angular resolution of $1^{\circ}$, and having $n$ channels as the time window history length.
We reproject past measurements as if they were captured from the robot's current position accounting for the robot's motion estimated by its odometry.
As a result, static obstacles yield overlapping points across the channels, while moving obstacles leave trail-like point patterns, as shown in Fig.~\ref{fig:history}.
The model is tasked to predict the presence of humans in the environment, and their distance and bearing relative to the robot.
For each ray, our model predicts: the scalar $\hat{p} \in [0, 1]$ representing the likelihood of human presence; the relative distance $\hat{d} \in [d_\text{min}, d_\text{max}]$, where $d_\text{min}$ and $d_\text{max}$ are the working range of the sensor; and the sine and cosine of the relative bearing $\hat{o} \in [-\pi, \pi]$ expressed as the difference between the person’s orientation and the direction parallel to the ray, with zero indicating a person directly facing the robot.
During inference, a discrete set of detections is obtained by thresholding and applying non-maxima suppression to the model's predicted presence.
We aim to extend the model's detection ability from the narrow area covered by the camera's \ac{fov} to the wider LiDAR's \ac{fov}, including areas where the supervision is scarce or absent, e.g. behind the robot.
To this end, we rely on the translational invariance of convolutions, in which patterns are detected regardless of their position within the input.
\begin{figure*}[t]
    \centering
    \begin{minipage}[t]{0.49\linewidth}
        \centering
        \frame{\includegraphics[width=\linewidth]{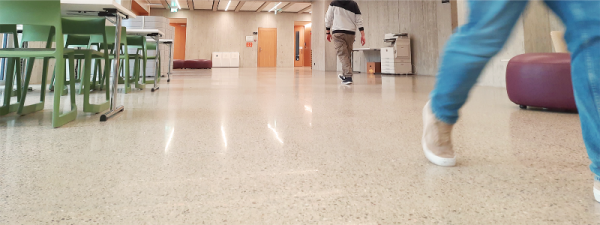}}\\[-5.5mm]
        \small {\color{black}\emph{University Corridor}}
    \end{minipage}
    \hfill
    \begin{minipage}[t]{0.49\linewidth}
        \centering
        \frame{\includegraphics[width=\linewidth]{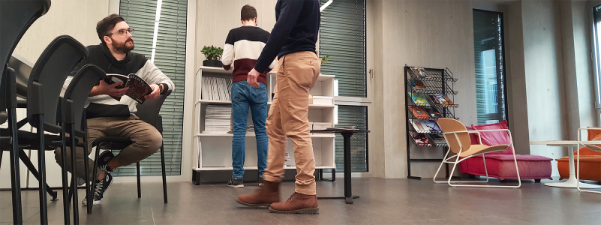}}\\[-5.5mm]
        {\color{white}\small \emph{Break Area}}
    \end{minipage}
    \\[1mm]
    \begin{minipage}[t]{0.49\linewidth}
        \centering
        \frame{\includegraphics[width=\linewidth]{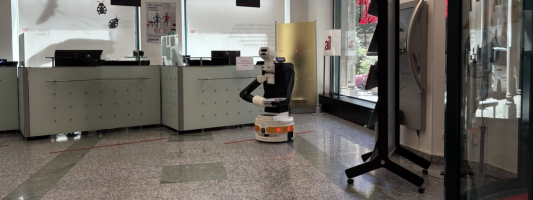}}\\[-5.5mm]
        {\color{white}\small \emph{City Office}}
    \end{minipage}
    \hfill
    \begin{minipage}[t]{0.49\linewidth}
        \centering
        \frame{\includegraphics[width=\linewidth]{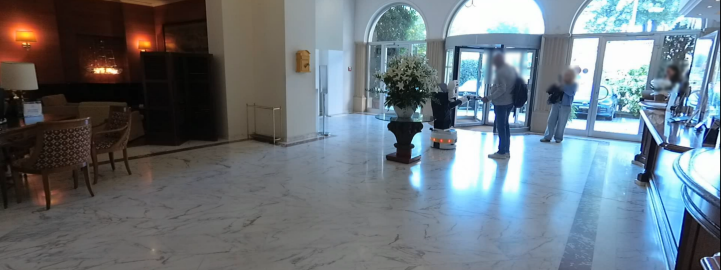}}\\[-5.5mm]
        {\color{white}\small \emph{Hotel Hall}}
    \end{minipage}

    \caption{Examples of environments used for training data collection (top row), and for in-the-wild experiments (bottom row).}
    \label{fig:scenarios}
    \vspace{-5mm}
\end{figure*}
The self-supervision signal is derived from the front-facing onboard camera providing the 3D pose of people's joints in the camera's \ac{fov}.
Among the joints, we specifically select the one located at the pelvis as it is closely tied to the legs' motion, and project its pose onto the horizontal plane to get 2D poses.
Labels for human presence $p$, distance $d$, and relative bearing $o$ are obtained from such poses for each ray of the LiDAR scan intersecting a person.
The presence $p$ is set to $1$ when a person is detected along the ray, $0$ otherwise; accordingly, the relative distance and bearing labels of people are assigned to rays in which they are detected, or left undefined otherwise.
Our model is trained to regress the three components using a masked loss, considering only errors generated from the rays corresponding to the area covered by the camera's \ac{fov}.
Additionally, distance and orientation losses are computed only for rays in which the supervision labels indicate the presence of humans.

\section{Experimental Setup}\label{sec:setup} 
\subsection{Hardware} 
We use a customized PAL Robotics TIAGo robot composed of a differential drive base, a prismatic torso, a $7$-axis arm, and a head that pans in the range $\pm75^{\circ}$ and tilts from $-60^{\circ}$ to $45^{\circ}$. 
Our TIAGo is equipped with additional sensors to better suit \ac{hri} applications: 
a Microsoft Azure Kinect RGB-D sensor mounted on the head tracking humans up to \SI{6}{m}~\cite{tolgyessy2021skeleton}, with a $65^{\circ}$ horizontal \ac{fov}, and a \SI{15}{Hz} frame rate;
%
a secondary LiDAR sensor mounted on the back of the robot's base, in addition to the built-in one located on the front, see Fig.~\ref{fig:task_and_sensor_setup}.
%
The front-facing LiDAR is at an height of \SI{95}{mm}, has a \ac{fov} of $190^{\circ}$, and scan rate of \SI{15}{Hz}; the back-facing one is at an height of \SI{329}{mm}, \ac{fov} of $255^{\circ}$, and scan rate of \SI{10}{Hz}; both sensors' operating range is \SI{0.05}-\SI{10}{m}.
To obtain a single, omnidirectional, and radially symmetric sensor, we fuse the two physical LiDARs into a \emph{virtual} one:
the two sensors' readings are time-synchronized at a rate of \SI{10}{Hz}, projected onto the 2D plane, expressed in the frame of the robot base, and aggregated into $1^\circ$-wide bins;
each bin is assigned the value of the closest point among its members; when there are no members, a default value of \SI{10}{m} is used.

\subsection{Dataset} 

\begin{figure*}[t!]
    \centering
    \frame{\includegraphics[trim={9cm 10cm 2cm 8cm},clip=true,height=0.25\textwidth]{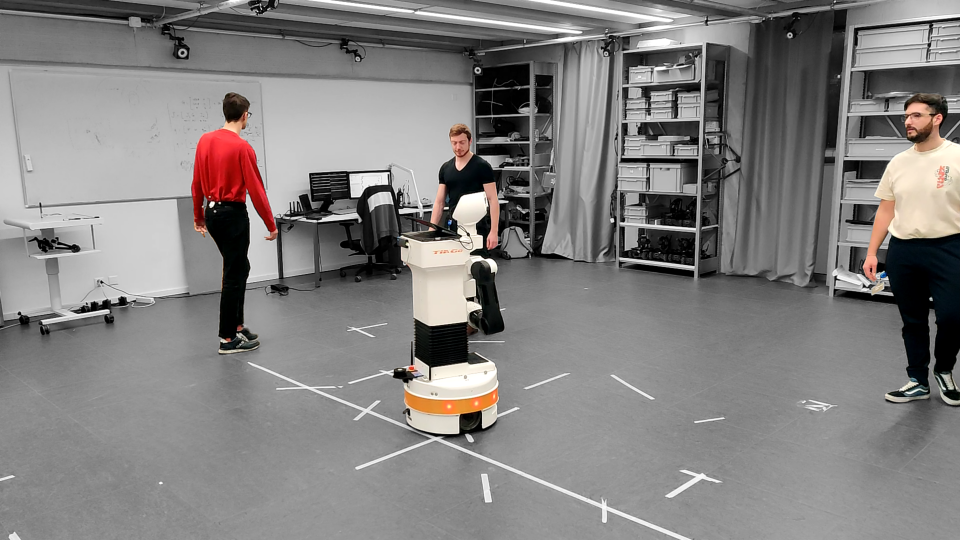}}
    \hfill
    \frame{\includegraphics[trim={2.0cm 6.5cm 2.0cm 7.0cm},clip=true,height=0.25\textwidth]{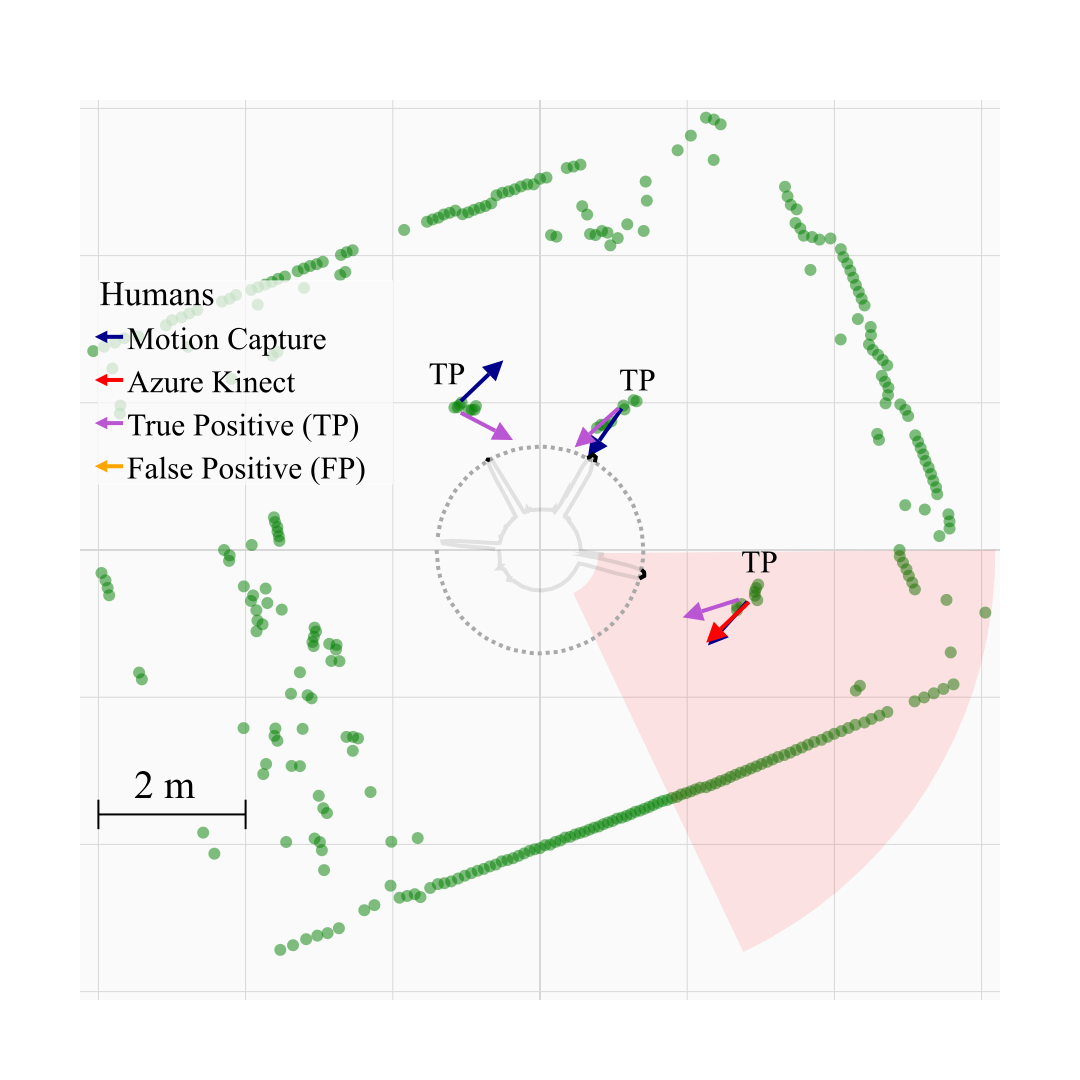}}
    \\[1.5mm]%
    \frame{\includegraphics[trim={9cm 10cm 2cm 8cm},clip=true,height=0.25\textwidth]{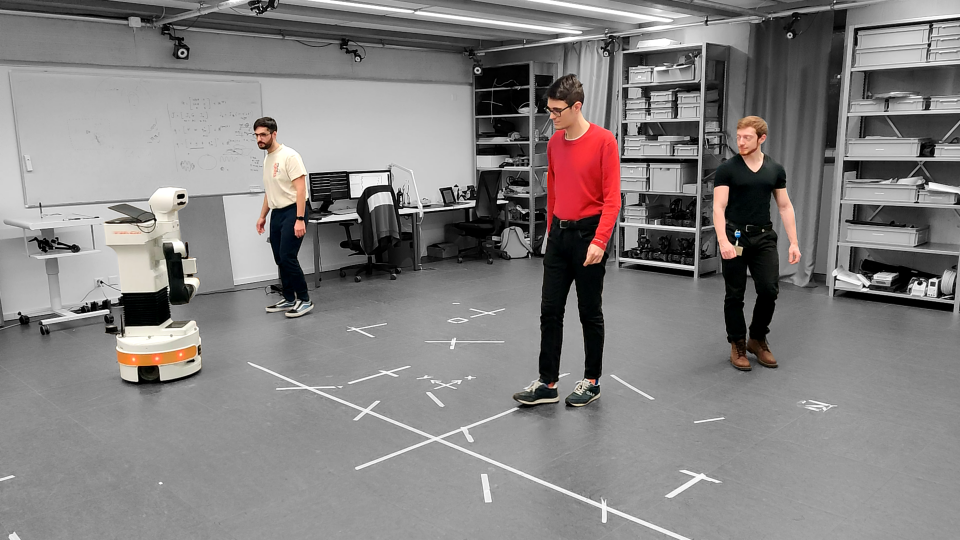}}%
    \hfill
    \frame{\includegraphics[trim={2.0cm 6.5cm 2.0cm 7.0cm},clip=true,height=0.25\textwidth]{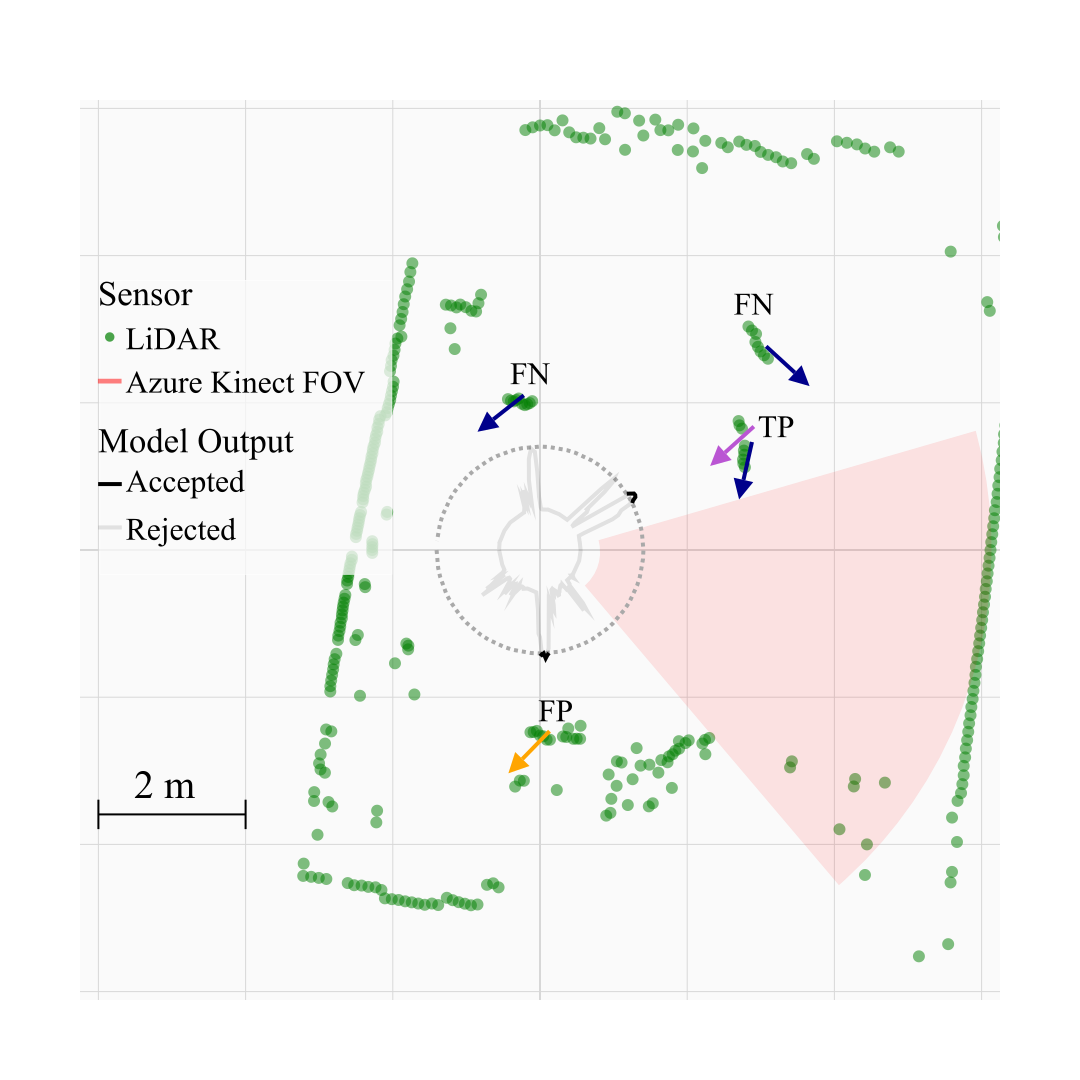}}%
    \caption{%
    Results on the test set: third-person view of two frames (left) and corresponding top view (right) depicting the {\color{lidar!75!black}LiDAR} scan; {\color{kinect}camera \ac{fov}} and detected pose arrows; {\color{optitrack}Motion Capture} ground truth pose arrows. The predicted presence $\hat{p}$ is shown as a {\color{wong_gray!75!black}gray line} when below the detection threshold of 90\% (dashed circle centered on the robot), or black otherwise. Predictions are represented by arrows colored differently for {\color{model_purple!75!black}true positives (TP)}, and {\color{model_yellow!75!black}false positives (FP)}.}
    \label{fig:lab_performance}%
    \vspace{-5mm}
\end{figure*}

We collected data across $9$ days in three environments (shown in Fig.~\ref{fig:scenarios} and~\ref{fig:lab_performance}): a public \emph{University Corridor} between classrooms with desks on the side and passers-by ($36$k samples); a large \emph{Break Area} with tables and chairs where expert individuals interact with the robot ($12$k samples); a \emph{Lab} setting with expert individuals interacting with the robot ($8$k samples).
During data collection, the robot's base motion and the head panning are randomized to increase the data variability and area covered by the camera. 
The robot base is manually controlled in the \emph{University Corridor} for security reasons whereas, in the other environments, it moves autonomously following random trajectories while avoiding collisions.
\begin{figure}[t]%
    \centering%
    \includegraphics[width=\columnwidth]{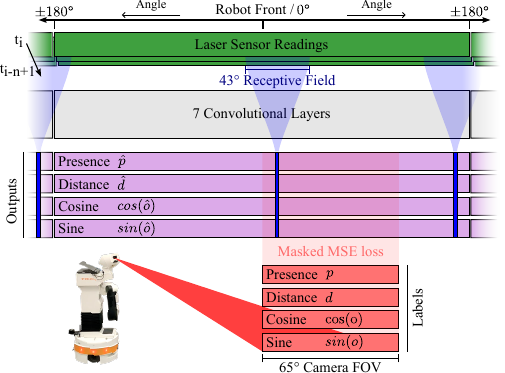}%
    \caption{Our model uses a temporal window of $n$ LiDAR scans to predict nearby people's presence $p$, distance $d$, and relative bearing~$o$ (represented by sine and cosine). Dilated circular convolutions handle omnidirectional scans and yield a $43^{\circ}$ {\color{model_purple!75!black}receptive field}. A masked MSE loss is only enforced on predictions that overlap with the {\color{kinect}camera \ac{fov}} (red shaded area).}%
    \label{fig:nn_architecture}%
    \vspace{-5mm}
\end{figure}
In all environments, we record body joints from the Azure Kinect and scans from the two LiDARs.
Additionally, the \emph{Lab} environment provides ground truth poses for people and the robot at \SI{100}{Hz} from an OptiTrack motion capture system featuring $18$ cameras.
The data is split into a train set composed of all samples from \emph{University Corridor} and half of those from \emph{Break Area}, totaling $42$k samples;
the remaining $6$k samples from \emph{Break Area} are used as validation set, and the $7$k samples from \emph{Lab} are used as the test set.
\subsection{Model architecture} 

Our model (see Fig.~\ref{fig:nn_architecture})  is composed of $7$ 1D convolutional layers with 32 output channels and layer normalization.
It features dilated circular convolutions with increasing kernel dimension from $3$ to $7$, resulting in a receptive field of $43^{\circ}$.
Dilation is used to have a low model complexity while achieving a receptive field large enough to effectively capture human motion.
During training, the model minimizes a squared error between predictions and label values of each output.
We train our model for 500 epochs at a constant learning rate of $3e^{-4}$ using the Adam optimizer~\cite{adam} and select the model weights resulting in the lowest validation loss.
We apply additive Gaussian noise and mirroring to the input as data augmentations during model training.

\section{Results}\label{sec:results}
\subsection{Quantitative performance in lab settings}
\begin{figure}[t]%
    \centering%
    \includegraphics[width=\linewidth]{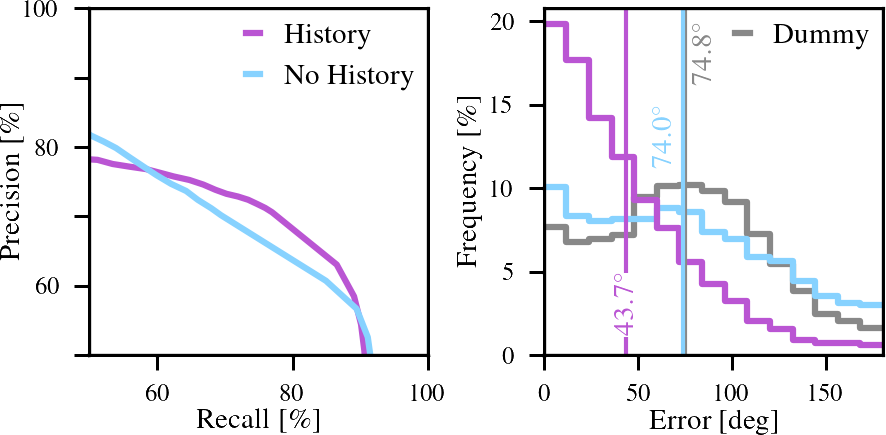}%
    \caption{Left: Precision-Recall curve for detection. Right: relative bearing error distribution. Results computed against mocap ground truth in the \emph{Lab} test set.}%
    \label{fig:plots}%
    \vspace{-5mm}
\end{figure}
We compare our model using the last $n=30$ scans collected at \SI{10}{Hz} over \SI{3}{s}, referred to as \emph{History}, with an ablated baseline named \emph{No History}, which uses only the current sensor reading as input ($n=1$).
All models are evaluated using the motion capture ground truth poses available only in \emph{Lab}, see Fig.~\ref{fig:task_and_sensor_setup}.
Figure~\ref{fig:plots} (left) reports the Precision--Recall curve illustrating the human detection performance of our models, obtained by progressively increasing the detection threshold applied to the presence output $\hat{p}$.
Predicted human positions are obtained by projecting a point in the direction of the corresponding ray at the estimated distance $\hat{d}$.
A prediction is considered a match (true positive) when its Euclidean distance from the ground truth is smaller than \SI{50}{cm}; the orientation component does not influence the matching criterion.
The plot shows that for recall values above $60\%$, the \emph{History} model (in purple) consistently outperforms \emph{No History} (in light blue).
Figure~\ref{fig:plots} (right) compares the orientation error distributions and include a \emph{Dummy} model that always returns the average ground-truth orientation and distance in the test set.
Errors are computed only for matched predictions (true positives), following the same procedure used for the Precision--Recall curve, while for the \emph{Dummy} model we consider the entire test set, assuming ideal detections.
Results indicate that temporal information is essential for accurate relative bearing estimation: the \emph{History} model achieves a mean absolute orientation error of $43^\circ$, compared to $74^\circ$ for \emph{No History} and $75^\circ$ for the \emph{Dummy} baseline.
For distance estimation, the \emph{History} model attains a mean absolute error of \SI{13}{cm}, slightly higher than the \SI{10}{cm} of \emph{No History} and the \SI{64}{cm} of \emph{Dummy}, suggesting that temporal context primarily benefits orientation estimation rather than range accuracy.
In terms of human detection, the \emph{History} model achieves a precision at $80\%$ recall ($P_{80}$) of $71.2\%$, outperforming the \emph{No History} baseline ($P_{80}=60.7\%$).
These results confirm that temporal context notably improves detection reliability and spatial awareness, aligning with prior work on human motion modeling~\cite{liu2013accurate, Luo_2020_ACCV}.
\begin{table}[b]
    \setlength\tabcolsep{1.2mm} 
    \renewcommand{\arraystretch}{1.2} 
    \centering
    \caption{Performance across the deployment environments.}
    \label{tab:table_public_envs}
    \begin{tabular}{lllrrr}
    \toprule
    \multirow{2}{*}{Ground Truth} & \multirow{2}{*}{Scenario} & Dimension & \multicolumn{1}{c}{$P_{80}$} & \multicolumn{1}{c}{$E_\text{o}$} & \multicolumn{1}{c}{$E_\text{d}$} \\[-2pt]
    & & [\# samples] & [\%] $\uparrow$ & [deg] $\downarrow$ & [cm] $\downarrow$ \\
    \midrule
    Motion Capture & \emph{Lab}          & 8k  & $71.2$ & $43$ & $13$ \\
    Azure Kinect   & \emph{Lab}          & 8k  & $76.4$ & $35$ & $12$ \\
    Azure Kinect   & \emph{City Office}  & 6k  & $71.5$ & $38$ & $16$ \\
    Azure Kinect   & \emph{Hotel Hall}   & 42k & $62.7$ & $41$ & $16$ \\
    \bottomrule
    \end{tabular}
    \vspace{-2mm}
\end{table}
\subsection{Deployment in public spaces}
\begin{figure}[t!]
    \centering
    \frame{\includegraphics[trim={0 11cm 0 8cm},clip=true,width=\columnwidth]{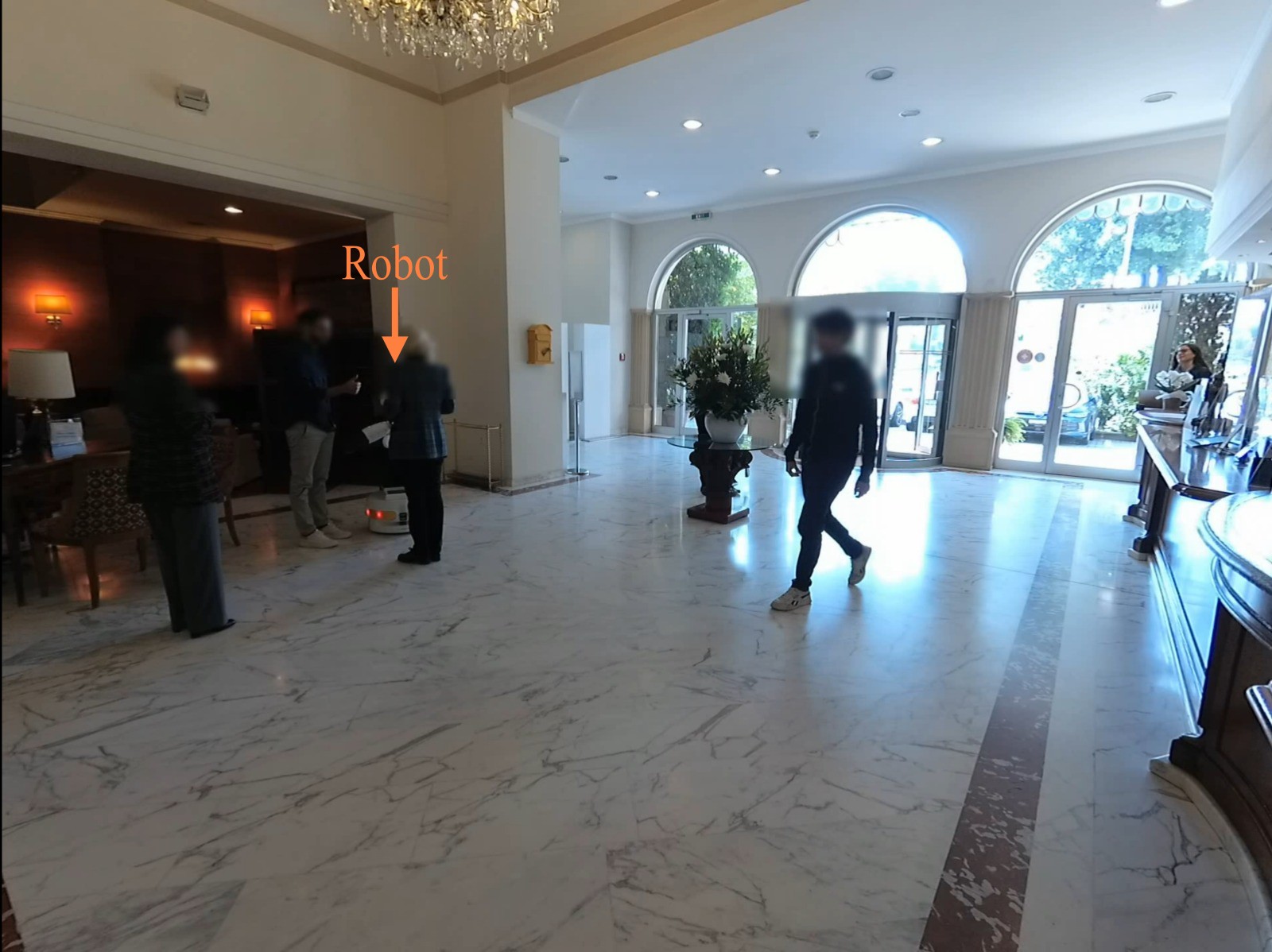}}
    \\[5pt]
    \frame{\includegraphics[trim={2cm 11cm 2cm 8cm},clip=true,width=\columnwidth]{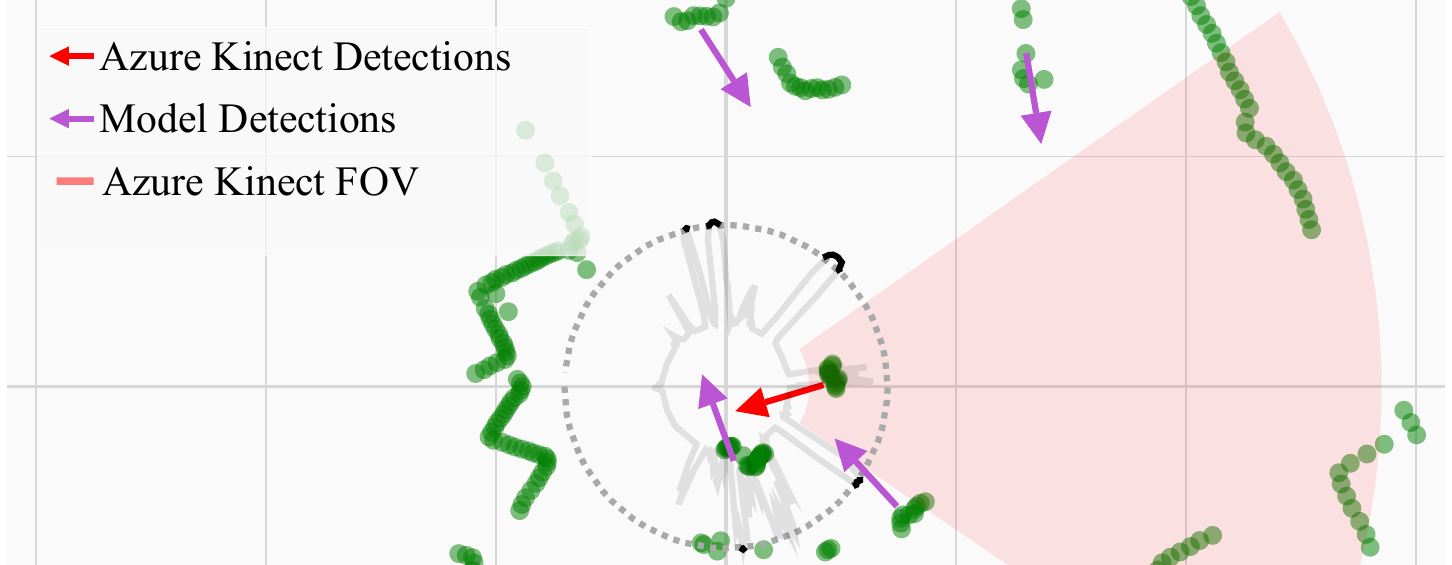}}
    \vspace{-5mm}
    \caption{Example of model performance deployed in the \emph{Hotel Hall} scenario.
    }
    \label{fig:hotel_performance}
    \vspace{-2mm}
\end{figure}

To further evaluate the generalization capability of our \emph{History} model, we deployed the robot in-the-wild, engaging with non-expert users in two public scenarios: the \emph{City Office}, a public administrative office, and the \emph{Hotel Hall}, both shown in Fig.~\ref{fig:scenarios}.
An snapshot of our system operating in the \emph{Hotel Hall} is shown in Fig.~\ref{fig:hotel_performance}.
In these environments, motion capture data are unavailable; therefore, we assess the model performance against the onboard Azure Kinect detections, albeit only within the camera's \ac{fov}.
The results, summarized in Tab.~\ref{tab:table_public_envs}, demonstrate that the learned representation generalizes well to unseen environments without any additional fine-tuning.
It is worth noting that, being a self-supervised approach, our method can be deployed in any environment and, by continuously collecting new data, can autonomously generate self-labels, retrain the detector, and adapt over time to changing environmental conditions.
From an HRI perspective, these results suggest that the model provides peripheral awareness around the robot, particularly in regions outside the camera field of view. This capability can support smoother transitions between navigation and interaction by enabling earlier detection of nearby people, more appropriate robot orientation, and more selective activation of head-mounted sensing. In social and service settings, such awareness is especially valuable when people approach from the side or behind the robot.
Qualitative results and examples of the system’s behavior in these real-world deployments are presented in the accompanying video available at \url{https://youtu.be/Wlw-mB84FPk}. 
\section{Conclusion}\label{sec:conclusions}
We presented a lightweight approach for human detection and pose estimation for service robots using 1D LiDAR sensors.
It leverages a state-of-the-art detector from an RGB-D camera as source of self-supervision, requiring no pre-collected datasets.
The approach can adapt to different sensing setups, assuming only a precise albeit narrow source of supervision for interpreting readings from a much wider FOV, possibly omnidirectional sensor.
The code to collect data, train models, and run them in real-time is made publicly available for the benefit of the community (\url{https://github.com/idsia-robotics/sixth_sense_lhpe}); we also provide our datasets (\url{https://zenodo.org/records/14936069}).
The approach has been validated in-the-wild, within relevant scenarios.
Compared with RGB-D or 3D LiDAR human detection and pose estimation approaches, the proposed method yields a coarser but more deployable pipeline using lower-cost and more accessible sensors. It could support commercial robots such as vacuum cleaners or lawn mowers by providing a lightweight estimate of human presence in the deployment environment, but it could also complement high-confidence, narrow-field-of-view perception on larger robotic platforms. This trade-off is particularly relevant for service robots operating in public and shared spaces, where cost, coverage, and privacy constraints often limit the sensing suite.
Future work will explore how this awareness layer can support HRI, e.g. engagement initiation, yielding, person selection for service interactions, and intention-to-interact estimation.

\bibliographystyle{IEEEtran}
\bibliography{references}

\end{document}